\documentclass{bmvc2k}

\usepackage{multirow, hhline}
\usepackage{amsmath}
\usepackage{xspace}
\usepackage{amssymb}
\usepackage{booktabs}

\usepackage{graphicx}
\usepackage{pgfplots} 
\usepackage{siunitx}
\usepackage[nolist,nohyperlinks,printonlyused]{acronym}

\DeclareMathOperator*{\argmax}{arg\,max}

\newcommand{\NViews}{K}

\def\etc{etc.\@\xspace}
\def\mathbi#1{\textbf{\em #1}}

\newcommand{\OurMethod}{ViLGOD\xspace}
\newcommand{\OurMethodPlus}{ViLGOD-CP\xspace}

\newcommand{\dbscan}{\text{DBSCAN}}

\newcommand{\figref}[1]{Fig.~\ref{#1}}
\newcommand{\secref}[1]{Section~\ref{#1}}

\newcommand{\tabref}[1]{Table~\ref{#1}}

\def\eg{\emph{e.g}\bmvaOneDot}
\def\ie{\emph{i.e}\bmvaOneDot}

\def\wrt{w.r.t.}

\newcommand{\wod}{WOD}

\title{Vision-Language Guidance for LiDAR-based Unsupervised 3D Object Detection}

\addauthor{Christian Fruhwirth-Reisinger}{reisinger@tugraz.at}{1,2}
\addauthor{Wei Lin}{lin@ml.jku.at}{3}
\addauthor{Dušan Malić}{dusan.malic@tugraz.at}{1,2}
\addauthor{Horst Bischof}{bischof@tugraz.at}{1}
\addauthor{Horst Possegger}{possegger@tugraz.at}{1,2}

\addinstitution{
	Christian Doppler Laboratory for\\Embedded Machine Learning
}

\addinstitution{
	Institute of Computer Graphics\\ and Vision\\
	Graz University of Technology
}
\addinstitution{
	Institute for Machine Learning\\
    Johannes Kepler University Linz
}

\runninghead{Fruhwirth-Reisinger, Lin, Malic, Bischof, Possegger}{ViLGOD}

\begin{document}

\maketitle

\begin{abstract}
Accurate 3D object detection in LiDAR point clouds is crucial for autonomous driving systems. 
To achieve state-of-the-art performance, the supervised training of detectors requires large amounts of human-annotated data, which is expensive to obtain and restricted to predefined object categories.
To mitigate manual labeling efforts, recent unsupervised object detection approaches generate class-agnostic pseudo-labels for moving objects, subsequently serving as supervision signal to bootstrap a detector.
Despite promising results, these approaches do not provide class labels or generalize well to static objects. 
Furthermore, they are mostly restricted to data containing multiple drives from the same scene or images from a precisely calibrated and synchronized camera setup.

To overcome these limitations, we propose a vision-language-guided unsupervised 3D detection approach that operates exclusively on LiDAR point clouds. 
We transfer CLIP knowledge to classify point clusters of static and moving objects, which we discover by exploiting the inherent spatio-temporal information of LiDAR point clouds for clustering, tracking, as well as box and label refinement. 
Our approach outperforms state-of-the-art unsupervised 3D object detectors on the Waymo Open Dataset~($+23~\text{AP}_{3D}$) and Argoverse 2~($+7.9~\text{AP}_{3D}$) and provides class labels not solely based on object size assumptions, marking a significant advancement in the field. Code will be available at \url{https://github.com/chreisinger/ViLGOD}.
\end{abstract}

\section{Introduction}
\label{sec:intro}
\begin{figure}[t]
    \begin{tabular}{cccccc}
    \includegraphics[width=.135\textwidth]{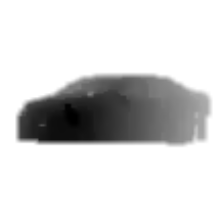} &   \includegraphics[width=.135\textwidth]{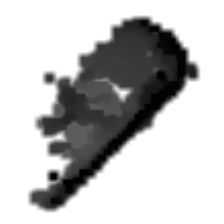} &   \includegraphics[width=.135\textwidth]{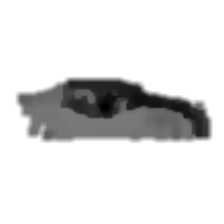} &
    \includegraphics[width=.135\textwidth]{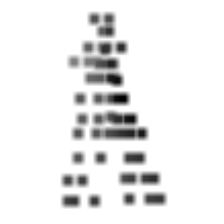} &   \includegraphics[width=.135\textwidth]{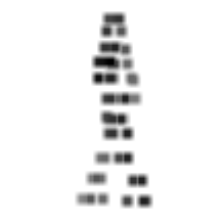} &   \includegraphics[width=.135\textwidth]{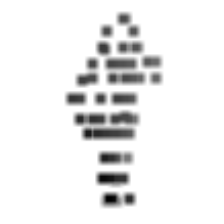}\\ 

    \includegraphics[width=.135\textwidth]{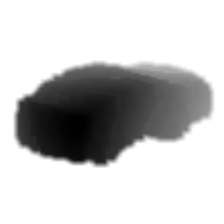} &   \includegraphics[width=.135\textwidth]{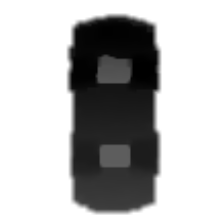} &   \includegraphics[width=.135\textwidth]{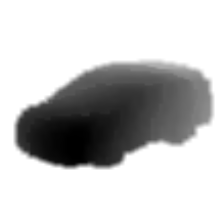} &
    \includegraphics[width=.135\textwidth]{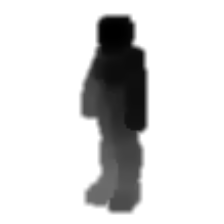} &   \includegraphics[width=.135\textwidth]{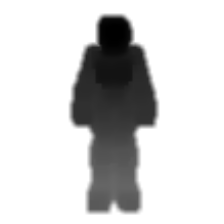} &   \includegraphics[width=.135\textwidth]{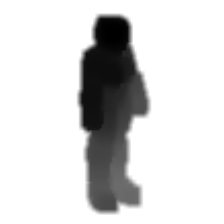}\\
    
    \end{tabular}
    \caption{\textbf{Comparison of point projections.}
            We illustrate two projection examples from LiDAR clusters of the WOD~\cite{sun_2020_waymo} (top row) and sampled CAD models~\cite{wu20153d} (bottom row) evaluated in~\cite{zhu_2022_pointclipv2}, in three different views. 
            While points sampled from CAD models produce consistently good results, LiDAR point cluster projections are negatively affected by incomplete clusters through self-occlusion (car, top left) and sparsity (pedestrian, top right).
            }
        \label{fig:teaser}
\end{figure}
For safe navigation and efficient path planning, autonomous vehicles critically rely on 3D object detection, \ie they must accurately identify the location, size, and type of objects (\eg~vehicle, cyclist, pedestrian) in the surrounding traffic environment.
Recent 3D object detectors~\cite{fan_2022_sst, lang_2019_pointpillars, shi_2019_pointrcnn, shi_2020_pvrcnn, shi_2021_pvrcnnpp, yan_2018_second} operate on the single modality of LiDAR point clouds~\cite{caesar_2020_nuscenes, geiger_2012_kitti} and require supervised training with vast amounts of manually annotated data, which is time-consuming and cost-intensive to obtain at a sufficient quality level.
Furthermore, despite their impressive performance, fully supervised 3D detectors lack the flexibility to cope with changing target data caused, for example, by different sensor setups~\cite{wang_2020_train} or unseen object classes. 
Re-annotation of this data would be necessary. 

Annotation-efficient solutions, such as semi-supervised approaches \cite{tang_2019_transferable3D, yin_2022_proficient_teachers, zakharov_2020_sdflabel} that require fewer manually labeled samples, or weakly-supervised methods employing techniques like click supervision~\cite{lee_2018_annotator, meng_2020_ws3d, meng_2022_w3d} already target these issues. 
However, these methods still require human interaction in the form of either hand-labeled data or a human-in-the-loop setup. 
Recent unsupervised 3D object detection approaches~\cite{najibi_2022_motioninspired, wang_2022_4d, you_2022_mobiledetection, luo_2023_reward, zhang_2023_oyster, baur_2024_liso} exhibit impressive performance in automatic labeling, requiring no prior knowledge other than the movement assumption and object size priors~\cite{luo_2023_reward}. 
However, such methods suffer from two major restrictions: 
First, they focus on localization and bounding box estimation, but do not provide class labels. 
Second, due to the lack of category information, they can merely discover moving objects, thus missing the detection of static foreground objects, which must be obtained in another fashion, \eg via repeated self-training~\cite{baur_2024_liso, zhang_2023_oyster}.
Furthermore, existing methods mostly require multiple drives from the same scene~\cite{luo_2023_reward, you_2022_mobiledetection}, demanding high-precision mapping equipment, additional camera images that are precisely calibrated and synchronized~\cite{wang_2022_4d}, or precise scene flow estimates~\cite{baur_2024_liso, najibi_2022_motioninspired}.
However, an unsupervised, class-aware detector using LiDAR scans from a single drive would be preferable for cost and performance reasons.

This paper addresses these issues by proposing \textbf{Vi}sion-\textbf{L}anguage \textbf{G}uidence for Unsupervised LiDAR-based 3D \textbf{O}bject \textbf{D}etection - \textbf{\OurMethod}.
Drawing inspiration from the recent success of vision-language foundation models, we employ CLIP~\cite{radford_2021_clip} to classify
objects in-the-wild. 
Specifically, we first propose spatio-temporal clustering over multiple frames incorporating motion cues to retrieve object proposals with high precision.
After filtering common-sense background samples, we project the remaining object proposal clusters into 2D image space to generate smooth depth maps from multiple views.
This simplification to the image domain allows us to obtain embeddings from the image path of the vision-language model.
By matching the visual embeddings against pre-processed text embeddings from the text path, we can acquire corresponding classification scores for the object proposals, independent of their movement status, resulting in zero-shot detection results.

The characteristics of LiDAR sensing, however, impose two unique challenges for projecting point clusters (see \figref{fig:teaser}) to leverage 2D vision-language models, which are not handled by existing approaches that deal only with CAD point clouds~\cite{zhang_2021_pointclip, zhu_2022_pointclipv2}:
1) LiDAR scans are 2.5D, \ie only the surface visible to the sensor is measured. This incomplete reconstruction restricts the variety in view points that is needed by \cite{zhang_2021_pointclip, zhu_2022_pointclipv2} to fully exploit the 2D visual embeddings.
2) LiDAR scans become increasingly sparse with larger distance to the sensor, making identifying the projected objects more and more difficult.
To mitigate these problems, we exploit the fact that LiDAR recordings are sequential. 
We design a simple but effective tracking and propagation module that allows the generation of different temporal views of the same object. This module enables, on the one hand, a more robust classification of objects and, on the other hand, the propagation of classes.
We further exploit the temporal object dependency to create bounding boxes and propagate them within tracks.

Our contributions are four-fold: 
(1)~\OurMethod is the first unsupervised but \emph{class-aware} 3D object detection method for outdoor LiDAR point clouds that provides class labels not solely based on object size heuristics;
(2)~\OurMethod operates on the single modality of LiDAR point clouds, and requires neither multiple drives throughout the same scene nor additional camera images;
(3)~In addition to moving objects, \OurMethod also localizes static objects through CLIP classification, thus provids valuable pseudo labels without the need for repeated self-training cycles.
(4)~Lastly, our detailed evaluations on the Waymo Open Dataset and Argoverse 2 demonstrate that even in the class-agnostic setup, \OurMethod outperforms the current state-of-the-art unsupervised 3D object detectors;

\section{Related Work}
\label{related_work}
\paragraph{Fully supervised LiDAR-based 3D object detection.}
Current state-of-the-art 3D object detection networks~\cite{fan_2022_sst, shi_2020_pvrcnn, shi_2021_pvrcnnpp, yan_2018_second, yin_2021_centerbased, zhou_2022_centerformer} typically rely on supervised learning methods and extensive quantities of human-annotated data~\cite{caesar_2020_nuscenes, mao_2021_once, sun_2020_waymo} to achieve peak performance. 
Depending on how they handle the sparse and unordered LiDAR point cloud input, these methods can be broadly divided into grid-based~\cite{fan_2022_sst, lang_2019_pointpillars, shi_2022_pillarnet, yan_2018_second, yin_2021_centerbased, zhou_2018_voxelnet, zhou_2022_centerformer}, point-based~\cite{shi_2019_pointrcnn, shi_2020_pointgnn, yang_2020_3dssd, zhang_2022_notallpoints}, and hybrid methods~\cite{shi_2020_pvrcnn, shi_2021_pvrcnnpp, yin_2021_centerbased}.
Instead of human annotations, all of these architectures can be trained with our high-quality pseudo-labels in a self-supervised fashion.

\paragraph{Label-efficient 3D object detection.}
Weakly-supervised methods learn from a limited amount of annotated data supplemented with auxiliary information, often by indirect supervision through image-level labels, coarse object locations, or scene-level annotations rather than 3D bounding boxes~\cite{huang_2022_representation, lee_2018_annotator, meng_2020_ws3d, meng_2022_w3d, tang_2019_transferable3D, zakharov_2020_sdflabel}. 
Semi-supervised methods, on the other hand, leverage a small amount of labeled data in conjunction with a large volume of unlabeled data~\cite{caine_2021_pseudo, qi_2021_offboard, yang_2021_auto4d, yin_2022_proficient_teachers}. 
Lastly, unsupervised methods strive to learn directly from the raw, unlabeled data, capitalizing on the inherent structure and distribution of the data and geometric properties. 
These methods frequently employ clustering techniques~\cite{caron_2020_swav, yin_2022_proposalcontrast}, contrastive learning~\cite{huang_2021_spatiotemporalself, liang_2021_exploringgeometry, liu_2023_fac, yin_2022_proposalcontrast} or masking~\cite{boulch_2022_also, krispel_2023_maeli, min_2022_voxelmae, xu_2023_mvjar} to derive meaningful representations from the data. 
Although all these methods have shown the potential to reduce the need for exhaustive manual annotations, they still require supervision in any form.

\paragraph{Unsupervised 3D object detection.}
Early methods for 3D object detection in LiDAR data~\cite{sualeh_2019_dynamic, zhang_2013_unsupervised3d, himmelsbach_2010_fastsegment} introduced the generic pipeline -- ground removal, clustering, bounding box fitting, and tracking -- which is the foundation for all recent unsupervised methods to acquire initial detections~\cite{dewan_2016_icra, najibi_2022_motioninspired, wang_2022_4d, you_2022_mobiledetection, zhang_2023_oyster, baur_2024_liso, seidenschwanz_2024_semoli, luo_2023_reward}. 
They then train deep neural networks with the initially generated pseudo labels to steadily improve the performance. 
However, existing methods are class-agnostic and thus lack the ability to find static objects in the initial label generation phase. Multiple rounds of self-training aim at mitigating this issue. For example, MODEST~\cite{you_2022_mobiledetection} and DRIFT~\cite{luo_2023_reward} leverage datasets with multiple drives of the same scene to detect moving objects. Recently, OYSTER~\cite{zhang_2023_oyster} leverages track consistency to find reliable pseudo-labels and introduces beam dropping for self-training to enhance detection quality in far ranges. Another line of work leverage 3D scene flow~\cite{najibi_2022_motioninspired, baur_2024_liso} and camera images~\cite{wang_2022_4d}.
In contrast, our \OurMethod already localizes and classifies both static and moving objects without training and does not require multiple drives or additional sensor modalities.

\paragraph{CLIP for 3D understanding.}
Explorations of transferring CLIP knowledge for 3D understanding start with basic tasks of 3D object classification in zero-shot~\cite{zhang_2021_pointclip,zhu_2022_pointclipv2,xue2022ulip,huang2022clip2point,zeng_2023_clip_square}, few-shot~\cite{zhang_2021_pointclip,zhu_2022_pointclipv2,huang2022clip2point,zeng_2023_clip_square} and fully-supervised settings~\cite{huang2022clip2point}. 
Recent approaches apply CLIP on non-trivial prediction tasks for 3D scene understanding in indoor environments~\cite{huang2022frozen, peng_2023_openscene, huang2023joint, ding2022language, yao20223d, lu_2023_openvocabulary}. 
However, only few approaches transfer CLIP for outdoor scene understanding on LiDAR point clouds:
\citet{peng_2023_openscene} align 3D point cloud features with corresponding camera images via a distillation loss and evaluate on outdoor open-vocabulary semantic segmentation. \citet{chen_2023_clip_2_scene} deploy an annotation-free semantic segmentation pipeline by enforcing consistency between point cloud features and the corresponding image features.
All these works rely on multi-modal inputs, utilizing the 2D images as a bridge to connect the point cloud and the language modality. 
In contrast, our method operates exclusively on LiDAR point clouds and exploits relatively simple techniques~\cite{zhang_2021_pointclip, zhu_2022_pointclipv2} for transferring CLIP knowledge to 3D data without requiring additional camera images.
However, careful adjustments are needed to deal with clustering errors, increasing sparsity for distant objects, and incomplete objects that all arise from 2.5D LiDAR scans.

\section{Vision-Language Guided 3D Object Detection}
\label{method}
\begin{figure*}[t!]
	\centering
	\includegraphics[width=.95\textwidth]{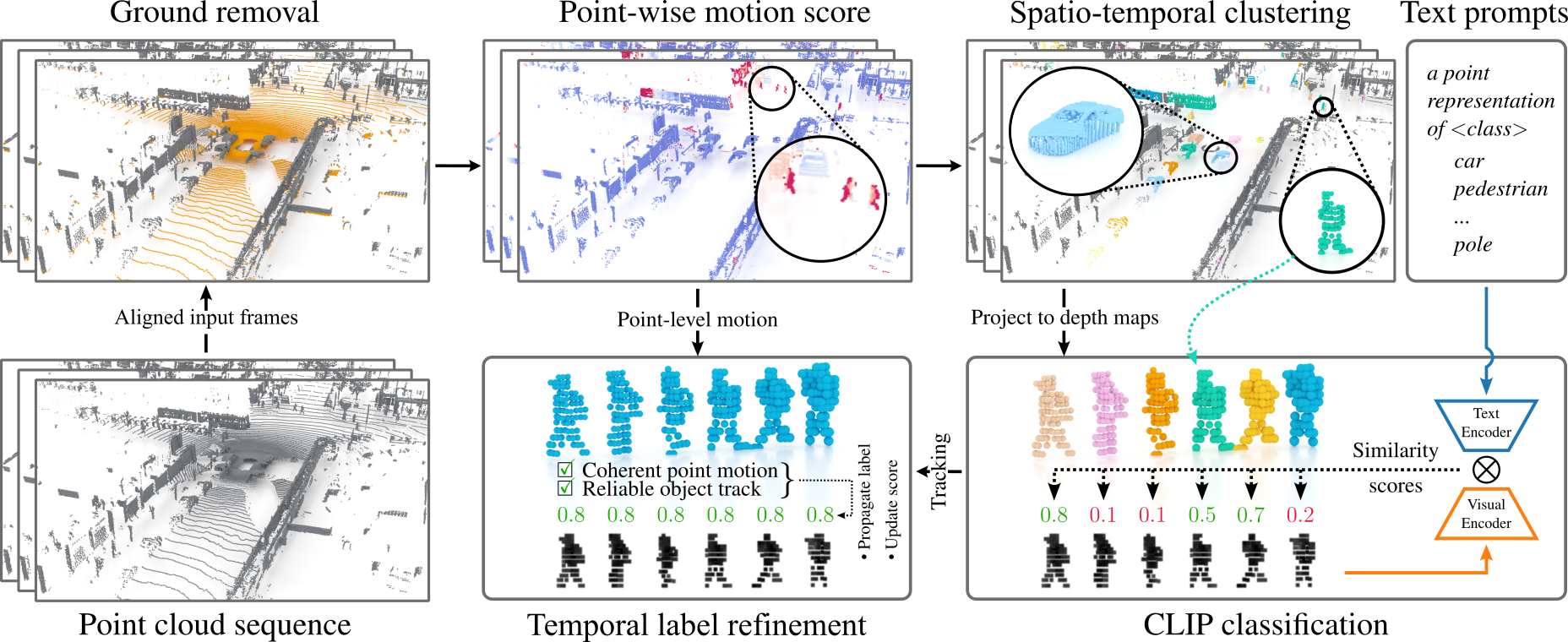}
	\caption{
		\textbf{\OurMethod overview.} After spatio-temporal clustering and filtering, we project 3D point clusters into 2D depth maps, subsequently fed to CLIP for zero-shot recognition. Objects close to the ego-vehicle result in smooth depth-maps, which can be correctly classified with high certainty, \eg the car in the bottom right. Distant objects, on the other hand, are more challenging and require additional context information to improve classification results, \eg the pedestrian on the top right. We omit bounding boxes to enhance clarity. 
	}
	\label{fig:overview}
\end{figure*}

We aim to detect 3D objects solely from LiDAR point clouds, without training on labeled data.
To realize this fully unsupervised, yet class-aware approach, we leverage the spatial and temporal cues inherently available in sequential LiDAR scans in combination with the powerful multimodal capabilities of recent vision-language models, as illustrated in Fig.~\ref{fig:overview}.
In particular, we first extract object proposals for both, moving and static objects, by spatio-temporal clustering, filtering and bounding box fitting (Section~\ref{sec:discovery}).
To obtain category estimates for these discovered proposals, we then employ the vanilla image-text foundation model CLIP to classify depth map projections in a multi-view aggregation setup (Section~\ref{sec:class-aware}).
In the final step, we leverage the temporal knowledge to refine and propagate bounding boxes and class labels throughout the LiDAR sequence, resulting in improved predictions even for distant objects.
We demonstrate that our training-free zero-shot detection results can be leveraged as pseudo labels for any supervised 3D detector (Section~\ref{sec:pseudo-label-training}). 

\subsection{Unsupervised Object Discovery}
\label{sec:discovery}
We denote a LiDAR point cloud sequence with $T$ frames as $\mathcal{P}=\{\mathcal{P}^t\}^T_{t=1}$, where the point cloud in the $t$-th frame is denoted as $\mathcal{P}^t = \{\mathbi{p}^t_i\}^{N^t}_{i=1}$, which represents a set of 3D points $\mathbi{p}_i^t \in \mathbb{R}^3$. 
In order to enable the unsupervised clustering of spatially related objects, we first remove ground points in each LiDAR scan.
Specifically, we perform ground segmentation with Patchwork++~\cite{lee_2022_patchworkpp} that applies plane fitting on concentric zone patches, and outputs a set of ground points $\mathcal{G}^t$. 
We apply RANSAC~\cite{fischler_1981_ransac} to fit a ground plane $\mathcal{O}_g$ with the ground points. 
To find moving objects, we derive the point-level motion information by identifying ephemeral points~\cite{barnes2018ephemeral} in consecutive frames with the persistence point scores (PP-score)~\cite{you_2022_mobiledetection, you_2022_udarepeatedtraversals}. 
The PP-score is computed as a measure of persistency for a 3D point by counting the number of its local neighbors across adjacent frames. 
Subsequently, we perform object discovery on non-ground points $\hat{\mathcal{P}}^t$ = $\mathcal{P}^t \setminus \mathcal{G}^t$. 

\paragraph{Proposal generation.} 
Given that 3D semantic entities consist of spatially related points, we follow previous work~\cite{zhang_2023_oyster} and apply HDBSCAN~\cite{campello_2013_hdbscan} for clustering.
However, to reduce over-segmentation and noise, we first transform $n$ frames of non-ground points $\hat{\mathcal{P}}^{t,\dots, t+n}$ into the reference frame $\hat{\mathcal{P}}^t$, sub-sample each frame by $1/n$ and concatenate the remaining points into a single frame $\hat{\mathcal{P}}^t_0$. In addition, we keep all points of frames where the PP-score indicates motion since moving points most likely belong to objects of interest. As input for clustering, we use the spatial position (x, y, z), the PP-score and the time difference $\Delta t$ w.r.t.~the reference frame for each point $\mathbi{p}_i^t$. The additional temporal input features help the clustering algorithm to distinguish 1)~moving from static objects and 2)~moving from moving objects which occupy the same space at different times (\eg crossing pedestrian trajectories). As a result, we receive cluster segments for $\hat{\mathcal{P}}^t$.
In a preliminary filtering step, we eliminate the most probable background objects. 
These are objects that either do not contain a minimum number of points, or are not located on the ground plane $\mathcal{O}_g$. 
This filtering results in a reduced set of segments $\mathcal{S}^t = \{\mathcal{S}_i^t\}^{M^t}_{i=1}$.
Afterwards, we follow~\cite{zhang_2017_lshape} and fit oriented 3D bounding boxes $\mathcal{B}^t = \{\mathbi{b}^t_i\}^{M^t}_{i=1}$ for all segments in $\mathcal{S}^t$. We denote the $i$-th bounding box as $\mathbi{b}^t_i = (b_x, b_y, b_z, b_l, b_w, b_h, b_\varphi)^\top \in \mathcal{B}^t$ where $b_x$, $b_y$, $b_z$ are the center coordinates, and $b_l$, $b_w$, $b_h$, $b_\varphi$ the length, width, height and orientation, respectively. 

\paragraph{Temporal coherence.} 
In order to distinguish between \textit{moving} and \textit{static} objects, we further exploit the temporal coherence in LiDAR sequences. 
First, we gather the point-level motion information within each segment $\mathcal{S}_i^t$ to determine the motion status of the corresponding bounding box $\mathbi{b}^t_i$. 
Specifically, if there is a percentile $\alpha$ of points within $\mathcal{S}_i^t$ which have PP-scores above a threshold $\delta$, we consider $\mathbi{b}^t_i$ \textit{static}, and \textit{moving} otherwise. 
Second, we perform multi-target tracking on all bounding boxes throughout the LiDAR sequence with greedy assignment.
Hence, a track is terminated when it has not been matched with any new incoming bounding box after a certain period, and each unassigned bounding box initiates a new track. 
We determine that a track and its bounding boxes are \textit{static} if 1)~all of its bounding boxes overlap with the largest box of the track and 2)~none of the bounding boxes was considered \textit{moving} according to the PP-scores. 
This distinction provides us groups of moving and static objects, which we assign class labels in the following.

\subsection{Vision-Language Guided Object Classification}\label{sec:class-aware}
\paragraph{CLIP preliminary.} 
CLIP~\cite{radford_2021_clip} is a large-scale vision-language model massively pre-trained via contrastive learning on 400M web image-text pairs, matching web images with their language descriptions. 
CLIP has a dual-encoder architecture comprising a visual encoder $\phi_v$ and a text encoder $\phi_t$. 
Given an input query image $x$ and a set of category text prompts $\mathcal T=\{t_c\}^{N_C}_{c=1}$, we denote the L2-normalized visual and text features as $z_v=\phi_v(x)$ and $\mathcal Z_t=\{  \phi_t(t_c)\}^{N_C}_{c=1}$. 
The zero-shot image classification is performed by selecting the class prompt with the maximum similarity to the visual representation, \ie $\hat c= \argmax_c \phi_v(x)^\top \phi_t(t_c)$. 

\paragraph{Transfer CLIP knowledge for 3D recognition.} 
A vision-language foundation model like CLIP cannot be directly applied for recognition tasks on 3D LiDAR point clouds. 
Therefore, we project the zero-centered 3D points within each bounding box into natural-looking 2D depth maps to mitigate the modality gap between unordered sparse point clouds and grid-based dense image pixels. 
Specifically, we follow the shape projection proposed for dense CAD point clouds \cite{zhu_2022_pointclipv2}, which consists of voxelization, densification, and smoothing, to projecting a 3D object instance into realistic depth maps. 
To preserve the 3D information, we generate the depth maps from multiple views after rotating and tilting the points in each bounding box. 
Examples of projected depth maps are illustrated in Figure~\ref{fig:teaser}.
We denote the number of views as $\NViews$, and the set of $\NViews$ depth maps projected from points in the $i$-th bounding box as $\mathcal X_i=\{x^k_i\}^{\NViews}_{k=1}$. 
For the category text prompts $t_c \in \mathcal T$, we use a 3D-specific prompt template \textit{a point representation of <class>}. 
Then, the zero-shot class label for the $k$-th view of the $i$-th object instance is $\hat c^k_i= \argmax_c \phi_v(x^k_i)^\top \phi_t(t_c)$.

\paragraph{Category text refinement.} For an improved zero-shot classification with CLIP, we refine the original category names.
Particularly, we replace the coarse category name \emph{vehicle} with a set of refined classes such as \emph{car}, \emph{truck}, \emph{bus}, and \emph{van}. 
Similarly, we replace the abstract category \emph{background} with instantiations of common non-traffic-participant objects such as \emph{traffic light}, \emph{traffic sign}, \emph{fence}, \emph{pole}, \etc
Finally, we add relevant synonyms, \eg \emph{human body} for \emph{pedestrian}. 
The detailed text refinement strategy is elaborated in the supplemental material.
After performing zero-shot classification on the expanded new category space, we merge the prediction results onto the fewer coarse classes in the original category space. 

\paragraph{Multi-view label voting.} To improve the prediction accuracy, \cite{zhu_2022_pointclipv2} proposed aggregating the weighted class predictions of all $\NViews$ views projected from a CAD point cloud.
Due to the different sensing characteristics, however, CLIP predictions for LiDAR-based projections vary largely depending on the view point.
To mitigate this, we vote for the mostly predicted class label $\hat c_i$ within $K$ views of an object $i$ and set $y_i$ to the mean prediction score of the $\hat{K}$ views with the same class label, \ie $y_i = \sum_{\hat{k}} y^{\hat{k}}_i / \hat{K}$, where $y^{\hat{k}}_i= \phi_v(x^{\hat{k}}_i)^\top \phi_t(t_{ \hat c_i  } )$. 
If the number of votes are equal, we assign the class label with the maximum mean score.

\paragraph{Temporally-coherent label refinement.} 
For unsupervised LiDAR segmentations, the projected depth maps suffer from degraded quality due to clustering errors, sparsity of distant objects, and incomplete objects of the 2.5D scans, as illustrated in the top-right of Figure~\ref{fig:overview}. This leads to erroneous recognition results, especially on distant or incomplete objects. 
To compensate for this, we leverage the multi-target tracking results from Section~\ref{sec:discovery} and apply a refinement strategy to propagate category labels and refined bounding box estimates throughout tracks of moving and static objects: 
For each track, we propagate the most confident CLIP label along the track if it is reliable w.r.t.~the temporal progression of the track.
A reliable class prediction means that it appears for at least 60 \% of the track.
We observed that CLIP prediction scores for smaller and less well-represented classes (\eg pedestrians, cyclists, and background classes) are generally lower than those of vehicles.
Thus, we propagate vehicle labels if the predicted score exceeds 0.5 and other labels if they exceed 0.3.
Since \emph{moving} objects are most certainly objects of interest in our traffic scenario, we aim to label all of these. If we can not obtain a reliable CLIP prediction, but the object is moving, we assign the class (vehicle, pedestrian, or cyclist) based on the observed object size.

Not only the correct classification of the object is important, but also its size and position.
To reliably estimate the bounding boxes even for occluded or incompletely observed objects, we apply a temporal refinement:
We first calculate the median box of the $M$ box candidates which contain the most cluster points within a track.
For \emph{static} object tracks, we propagate this box estimate at the median position and obtain the orientation as the majority vote among the $M$ boxes.
For \emph{moving} object tracks, we follow OYSTER~\cite{zhang_2023_oyster} and propagate the box along the tracking direction, aligning the box not with the center but with the closest corner to the ego-vehicle.

\subsection{Self-training}\label{sec:pseudo-label-training}
Our training-free unsupervised detection approach provides high quality pseudo-labels for the supervised training of any arbitrary 3D object detection architecture. 
We demonstrate this by leveraging the unsupervised detection results as pseudo ground truth in a supervised learning setting without bells and whistles. 
In particular, we train Centerpoint~\cite{yin_2021_centerbased} with our pseudo-labels in a supervised and class-aware setup.  
We neither do multiple rounds of training and refinement~\cite{you_2022_mobiledetection, luo_2023_reward, baur_2024_liso} nor do we require additional augmentations~\cite{zhang_2023_oyster}.

\section{Experiments}
\label{experiments}
To demonstrate the capabilities of our \OurMethod, we conduct experiments on two large-scale LiDAR datasets.
First, we compare our method to state-of-the-art unsupervised object detectors in a \emph{class-agnostic} setup, where we merge all predicted foreground objects into a single class.
Second, we compare our \emph{class-aware} results to class-agnostic approaches with assigned ground truth labels. Finally, we provide a detailed ablation on the separate components of our \OurMethod.

\paragraph{Datasets.}
We conduct our evaluations on the challenging Waymo Open Dataset~(WOD)~\cite{sun_2020_waymo} and Argoverse~2~\cite{wilson_2021_argoverse2}. 
WOD contains $1000$ publicly available sequences with approximately $200$ frames each.
It is separated into $798$ training and $202$ validation sequences. 
We follow the evaluation protocol of~\cite{najibi_2022_motioninspired, baur_2024_liso}, \ie evaluating the area of $100\si{m} \times 40\si{m}$ around the ego vehicle and reporting average precision (AP) with an intersection over union (IoU) threshold of 0.4 in 3D and BEV. 
Following \cite{seidenschwanz_2024_semoli,baur_2024_liso}, objects that move faster than 1\si{\m/\s} are considered \textit{moving}.
Full-range evaluations and additional APH (Average precision and heading) scores are included in the supplemental material.
Argoverse~2 contains $700$ training and $150$ validation sequences with approximately 150 frames each.
We follow the evaluation protocol of~\cite{baur_2024_liso} and evaluate the area of $100\si{m} \times 100\si{m}$ around the ego vehicle and report AP with an IoU threshold of 0.3 in BEV.
For the sake of comparability, we merge objects with the ability to move into the single class \emph{movable}. 
In WOD, this affects all relevant object classes; in Argoverse~2, we exclude, for example, \emph{Barrier}, \emph{Traffic cone}, but also \emph{bicycle} since the object is not moveable without a rider (\ie the separate \emph{cyclist} class). 

\paragraph{Implementation Details.}
We build our detection pipeline on top of the OpenPCDet~\cite{pcdetteam_2020_openpcdet} framework (v0.6.0) and conduct all experiments with the provided base models. We utilize Centerpoint~\cite{yin_2021_centerbased} for the supervised pseudo-label training. 
For these experiments, we follow the standard protocol of OpenPCDet and optimize with Adam~\cite{diderik_2015_adam} in a one-cycle policy~\cite{smith_2018_onecycle} with a maximum learning rate of $0.003$. 
However, we train for only 10 epochs on 50\% of the training data and do not sample from a pseudo-label database.
We ran all our experiments on 4 NVIDIA\textsuperscript{\textregistered} RTX\textsuperscript{\texttrademark} A6000 GPUs. Further implementation details and parameters can be found in the supplemental material.

\begin{table}[t]
    \begin{center}
    \small
        \begin{tabular}{@{}llccccccccc@{}}
        \toprule
        \multicolumn{1}{c}{} &
        \multicolumn{1}{c}{Dataset:} &
        \multicolumn{6}{c}{\textbf{Waymo Open Dataset (WOD)}} &
        \multicolumn{1}{c}{} &
        \multicolumn{2}{c}{\textbf{Argoverse 2}}
        \\ \cmidrule(l){3-8} \cmidrule{10-11}
        \multicolumn{1}{c}{} &
        \multicolumn{1}{c}{Motion Category:} &
        \multicolumn{2}{c}{Movable} &
        \multicolumn{2}{c}{Moving} &
        \multicolumn{2}{c}{Static} &
        \multicolumn{1}{c}{} &
        \multicolumn{2}{c}{Movable}
        \\ \cmidrule(l){3-8} \cmidrule{10-11}
        &
        \multicolumn{1}{c}{Average Precision:} &
        \multicolumn{1}{c}{BEV} &
        \multicolumn{1}{c}{3D} &
        \multicolumn{1}{c}{BEV} &
        \multicolumn{1}{c}{3D} &
        \multicolumn{1}{c}{BEV} &
        \multicolumn{1}{c}{3D} &
        \multicolumn{1}{c}{} &
        \multicolumn{1}{c}{BEV} &
        \multicolumn{1}{c}{3D}
        \\ \cmidrule{1-8} \cmidrule{10-11}

        \parbox[t]{1mm}{\multirow{2}{*}{\rotatebox[origin=c]{90}{\scriptsize\textcolor{gray}{Unsupervised~~~}}}} 
        & \dbscan~\cite{campello_2013_hdbscan} & 0.027 & 0.008 & 0.009 & 0.000 & 0.027 & 0.006 & & 0.054 & 0.020\\ 
        
        & RSF \cite{deng_2023_rsf} & 0.030 & 0.020 & 0.080 & 0.055 & 0.000 & 0.000 & & \underline{0.074} & \underline{0.055} \\ 
        & SeMoLi \cite{seidenschwanz_2024_semoli} $\dagger$ & - & 0.195 & - & \textbf{0.575} & - & - & & - & - \\
        & LISO-CP~\cite{baur_2024_liso} & \underline{0.292} & \underline{0.211}  & \underline{0.272} & 0.204  & \underline{0.208} & \underline{0.140} & & - & - \\ 
        & \OurMethod$\ddagger$ & \textbf{0.363} & \textbf{0.323} & \textbf{0.280} & \underline{0.240} & \textbf{0.327} & \textbf{0.311} & & \textbf{0.251} & \textbf{0.225}\\ \cmidrule{1-8} \cmidrule{10-11}
        \parbox[t]{2mm}{\multirow{2}{*}{\rotatebox[origin=c]{90}{\scriptsize\textcolor{gray}{Self-train}}}} 
        & OYSTER-CP~\cite{zhang_2023_oyster} & 0.217 & 0.084 & 0.151 & 0.062 & 0.176 & 0.056 & & 0.381 & 0.150\\ %
        & LISO-CP~\cite{baur_2024_liso} & \underline{0.380} &  \underline{0.308} & \underline{0.350} & \underline{0.296} & \underline{0.322} & \underline{0.255} & & \underline{0.448} & \underline{0.367} \\ 
        & \OurMethodPlus$\ddagger$ & \textbf{0.564} & \textbf{0.538} & \textbf{0.540} & \textbf{0.521} & \textbf{0.465} & \textbf{0.454} & & \textbf{0.464} & \textbf{0.446}\\
        \bottomrule
        \end{tabular}%
    \end{center}
    \caption{
        \textbf{Class-agnostic evaluation} following the protocols of \cite{baur_2024_liso, seidenschwanz_2024_semoli} for \wod~\cite{sun_2020_waymo} (\ie AP for BEV and 3D, difficulty level L2, IoU 0.4) and \cite{baur_2024_liso} for Argoverse~2 \cite{wilson_2021_argoverse2} (\ie AP for BEV and 3D, IoU 0.3).
        Results for DBSCAN, RSF, OYSTER-CP taken from~\cite{baur_2024_liso}, $\dagger$:~Results from~\cite{seidenschwanz_2024_semoli}. $\ddagger$: Method uses CLIP, unsupervised pre-trained on text-image pairs.
    }
    \label{tab:waymo-class-agnostic}
\end{table}

\begin{table}[t]
    \begin{center}
    \small
        \begin{tabular}{@{}llcccccc@{}}
        \toprule
        \multicolumn{1}{c}{} &
        \multicolumn{1}{c}{Object Class:} &
        \multicolumn{2}{c}{Vehicle} &
        \multicolumn{2}{c}{Pedestrian} &
        \multicolumn{2}{c}{Cyclist} \\  
        \multicolumn{2}{c}{Average Precision:} &
        \multicolumn{1}{c}{BEV} &
        \multicolumn{1}{c}{3D} &
        \multicolumn{1}{c}{BEV} &
        \multicolumn{1}{c}{3D} &
        \multicolumn{1}{c}{BEV} &
        \multicolumn{1}{c}{3D} \\
        \midrule

        \multicolumn{8}{c}{\textit{GT class labels}} \\
        \cmidrule(l){2-8} 
        \parbox[t]{1mm}{\multirow{2}{*}{\rotatebox[origin=c]{90}{\textcolor{gray}{Unsupervised}}}} 
        & \dbscan~\cite{campello_2013_hdbscan} & 0.184 & 0.048 & 0.002 & 0.000 & 0.001 & 0.000 \\ 
        
        & RSF \cite{deng_2023_rsf} & 0.109 & 0.074 & 0.000 & 0.000 & 0.002 & 0.000  \\ 
        & LISO-CP~\cite{baur_2024_liso} & \textbf{0.607} & \underline{0.440} & \underline{0.029} & \underline{0.009} & \underline{0.010} & \underline{0.004}  \\ 
        \cmidrule(l){2-8} 
        \multicolumn{8}{c}{\textit{Predicted class labels}} \\
        \cmidrule(l){2-8} 
        
        & \OurMethod & \underline{0.490} & \textbf{0.448} & \textbf{0.168} & \textbf{0.141} & \textbf{0.076} & \textbf{0.074}  \\
        \midrule 
        \multicolumn{8}{c}{\textit{GT class labels}} \\
        \cmidrule(l){2-8} 
        \parbox[t]{2mm}{\multirow{2}{*}{\rotatebox[origin=c]{90}{\textcolor{gray}{Self-train}}}} 
        & OYSTER-CP~\cite{zhang_2023_oyster} & 0.562 & 0.204 & 0.000 & 0.000 & 0.000 & 0.000 \\ %
        & LISO-CP~\cite{baur_2024_liso} & \textbf{0.695} & \underline{0.543} & \underline{0.055} & \underline{0.037} & \underline{0.022} & \underline{0.016} \\ 
        \cmidrule(l){2-8} 
        \multicolumn{8}{c}{\textit{Predicted class labels}} \\
        \cmidrule(l){2-8} 
        & \OurMethodPlus$\ddagger$ & \underline{0.644} & \textbf{0. 624} & \textbf{0.388} & \textbf{0.359} & \textbf{0.075} & \textbf{0.075}\\
        \bottomrule
        \end{tabular}%
    
    \end{center}
    \caption{
         \textbf{Class-aware evaluation on \wod} \cite{sun_2020_waymo} following the protocol of \cite{baur_2024_liso} (\ie AP scores for BEV and 3D, difficulty level L2, IoU 0.4).
        Results for DBSCAN, RSF, OYSTER-CP taken from~\cite{baur_2024_liso}. $\ddagger$: Method uses CLIP, unsupervised pre-trained on text-image pairs.
    }
    \label{tab:waymo-class-aware}
\end{table}

\paragraph{Class-agnostic Results.}
The unsupervised 3D object detection results on the \wod~validation set are shown in Table~\ref{tab:waymo-class-agnostic}.
The direct comparison among all unsupervised methods shows their object discovery capabilities:
Our vision-language guidance allows \OurMethod to locate both moving \emph{and} static objects in a single pass without requiring any re-training cycles.
By leveraging the temporal coherence, we are also able to obtain accurate 3D bounding box estimates as indicated by the small gap between AP BEV to AP 3D.
Thus, our \OurMethod excels in retrieving object candidates that can be used as pseudo labels to train a detector.
To demonstrate this, we use these object proposals to train a Centerpoint~\cite{yin_2021_centerbased} detector from scratch (denoted \OurMethod-CP).
The results for this \emph{self-training} in Table~\ref{tab:waymo-class-agnostic} show that our object proposals lead to a substantially improved detection performance, despite training Centerpoint for only 10 epochs (without augmenting samples from the pseudo-label database).

\paragraph{Class-aware Results.}
Table~\ref{tab:waymo-class-aware} shows the results for our zero-shot detections (\ie class-aware predictions) in comparison to existing class-agnostic approaches with assigned ground truth (GT) labels.
The consistently high AP 3D scores show that our \OurMethod provides accurate object proposals that are well suited for training a detector.
In particular, our \OurMethod enables, for the first time, the training of a \emph{class-aware} detector in an efficient manner: Without any manual human intervention and without time-consuming repeated self-training cycles.
Notably, our approach leads to remarkable improvements in detecting the vulnerable road user classes (\ie pedestrians and cyclists). 

\paragraph{Ablation Study.} We conduct a detailed ablation study to show the contribution of each step of our approach. \tabref{tab:ablation_main} lists the zero-shot detection results (pseudo-labels) on the WOD validation set.  
In addition to class-agnostic scores, we provide class-aware results, as our method provides zero-shot class-label predictions, allowing for better analysis. 

The baseline is a simple combination of spatio-temporal clustering, CLIP~\cite{radford_2021_clip} classification, and L-shape bounding box fitting~\cite{zhang_2017_lshape}. 
As shown by the results, all steps contribute to the effectiveness of our unsupervised detection approach \OurMethod, allowing us to surpass the current state-of-the-art in unsupervised (class-agnostic) 3D object detection.
The simple preliminary filtering not only increases the performance by $2.3~\text{AP}_{BEV}$~/~$0.7~\text{AP}_{3D}$ but also speeds up the entire detection process by significantly reducing the number of remaining cluster segments. 
The corner-aligned box fitting~\cite{zhang_2023_oyster} for \emph{moving} objects particularly affects the detection quality of pedestrians. 
With an overall increase in performance by $2.9~\text{AP}_{BEV}$~/~$1.5~\text{AP}_{3D}$ for \emph{movable} objects, pedestrians benefit the most with an increase of $63.3\%$ from $4.9~\text{AP}_{BEV}$ to $8.0~\text{AP}_{BEV}$.
The most significant step, however, is the class label refinement. 
It increases the performance for all object classes, doubling the performance on \emph{pedestrians}, and enables the detection of at least some \emph{cyclists}.
Finally, the propagation of adjusted bounding boxes is especially advantageous for \emph{vehicles}. 
Incorporating this final refinement step improves the overall score by $6.3~\text{AP}_{BEV}$~/~$7.1~\text{AP}_{3D}$ and thus represents the most significant gain in absolute terms. Note that this improvement stems primarily from the \emph{vehicle} class, representing the largest proportion of objects present in the WOD.
\begin{table}[t]
	\small
    \begin{center}
        \begin{tabular}{lcccccc}
        \toprule
            & & \multicolumn{2}{c}{Movable} & Vehicle & Pedestrian & Cyclist \\
            & & BEV & 3D & BEV & BEV & BEV \\
            \midrule
            Baseline & & 0.199 & 0.183 & 0.320 & 0.046 & 0.013 \\
            \midrule
            +~Filtering &  & 0.222 & 0.190 & 0.349 & 0.049 & 0.016 \\
            +~Corner alignment &  & 0.251 & 0.205 & 0.359 & 0.080 & 0.014 \\
            +~Class label refinement & & 0.301 & 0.252 & 0.390 & 0.163 & 0.064\\
            +~Bounding box refinement &  & 0.363 & 0.323 & 0.490 & 0.168 & 0.076\\
            
        \bottomrule
        \end{tabular}
    \end{center}
	
	\caption{
        \textbf{Ablation study} following the protocols of~\cite{najibi_2022_motioninspired, baur_2024_liso} on the WOD~\cite{sun_2020_waymo} (\ie AP for BEV and 3D, difficulty level L2, IoU 0.4). The baseline includes spatio-temporal clustering, CLIP~\cite{radford_2021_clip} classification, and L-shape bounding box fitting~\cite{zhang_2017_lshape}. 
        The ablations are split into filtering and temporally coherent refinement steps (\ie, corner alignment, class label refinement, bouncing box refinement).
	}
	\label{tab:ablation_main}
\end{table}

\section{Conclusion}
We proposed \OurMethod, the first fully unsupervised, yet class-aware 3D object detection method for LiDAR data.
We combine the strong representation capabilities of vision-language models with unsupervised object discovery for both static and moving objects.
This enables zero-shot detections, which result in reliable pseudo labels when propagated throughout LiDAR sequences.
These pseudo-labels can be directly utilized to train a 3D object detector in a supervised manner, without the need for multiple self-training iterations.
Our evaluations demonstrate the potential of this fully unsupervised data exploration strategy to significantly reduce the manual annotation costs needed to obtain sufficient amounts of data to train current state-of-the-art detectors.

\label{conclusion}

\section*{Acknowledgements}
The financial support by the Austrian Federal Ministry for Digital and Economic Affairs, the National Foundation for Research, Technology and Development and the Christian Doppler Research Association is gratefully acknowledged. 

\appendix

\section{Implementation Details}
\label{sec:implementation}
\paragraph{Object discovery.} For object segmentation with HDBSCAN~\cite{campello_2013_hdbscan} we use $min\_cluster\_size=15$, $min\_samples=15$ and $cluster\_selection\_epsilon=0.15$. 
To avoid uncertain objects,~\ie very small objects that lead to ambiguous 2D depth map projections and flying objects, which are mostly likely no objects of interest caused by occlusion, we apply filters after clustering: We remove segments with less than $10$ points, exceeding the distance of $1\si{m}$ to the ground plane and segments with a height below $0.5\si{m}$ resulting in segments $\mathcal{S}_i^t$.

A track is considered \emph{static} when a percentile $\alpha=20\%$ of PP-Scores per segment $\mathcal{S}_i^t$ is above the threshold $\delta=0.7$ for all its segments, all the boxes of a track overlap with the largest box of the track, or the track has no consistent motion behavior (no smooth linear motion). 
We limit the greedy assignment between track predictions and detections by a $1\si{m}$ radius~\wrt~Euclidean distance. 
If no assignment can be found, we relax the radius to $5\si{m}$ and assign a matched detection if the number of points between the track prediction and detection cluster segment differs less than $30\%$. 
This relaxed assignment recovers very fast-moving objects, such as vehicles on the highway, and mitigates false assignments between temporally occluded or over-segmented objects.

\paragraph{Object classification.}
We use CLIP\footnote{CLIP model \href{https://openaipublic.azureedge.net/clip/models/5806e77cd80f8b59890b7e101eabd078d9fb84e6937f9e85e4ecb61988df416f/ViT-B-16.pt}{URL}}~\cite{radford_2021_clip} with the ViT-B/16 visual encoder~\cite{dosovitskiy_2021_vitb} for the classification of the projected depth maps. 
Therefore, we generate $K=4$ different views, containing the basic view without rotation, rotations about the z-axis (yaw) of $\pm 18^{\circ}$ and about the y-axis (pitch) of $6^{\circ}$. 
Compared to synthetic CAD data~\cite{zhu_2022_pointclipv2}, objects in outdoor LiDAR scans suffer from self-occlusion (recall Fig. 1 in the main manuscript) and thus are only sufficiently visible from small variations of the original viewpoint. 
Therefore, we only apply small rotations. 
A tilt in the negative y-direction is also not useful because the ground plane prevents a valid shape at the bottom (in contrast to, for example, viewing the roof of a car from a slightly elevated viewpoint).
In \tabref{tab:synonyms}, we provide the refined object categories we use for CLIP classification. 
The class predictions of the refined categories are later mapped back to the original object classes. 

\begin{table}[h!]
    \small
    \begin{center}
        \begin{tabular}{lc}
            \toprule
            Class & Refined categories \\
            \midrule
            Vehicle & car, truck, bus, van, minivan, pickup truck, school bus, fire truck, ambulance \\
            Pedestrian & pedestrian, human body, human \\
            Cyclist & cyclist, rider, bicycle, bike \\
            Background & traffic light, traffic sign, fence, pole, clutter, tree, house, wall \\
            \bottomrule
        \end{tabular}
    \end{center}
	\caption{
        \textbf{Category text refinement}. We use the listed refined categories for predicting class labels with CLIP and map the result back to the original class space of the dataset.
	}
	\label{tab:synonyms}
\end{table}

\paragraph{Temporally-coherent class label refinement.} After classification, we assign the class label with the highest score to all objects in the track as long as the maximum class label score exceeds the threshold of 0.5 for \emph{vehicles} and 0.3 for \emph{pedestrians}, \emph{cyclists} and \emph{background}. Additionally, this class label must match at least $60\%$ of the tracks' predicted classes. 
This propagation of labels throughout track is done for \emph{static} and \emph{moving} objects. We keep the CLIP label prediction for \emph{static} objects not fulfilling the proposed conditions.

However, assuming that all objects in motion are of interest and the class label space in the automotive domain contains \emph{vehicles}, \emph{pedestrians} and \emph{cyclists}, we added a default classification scheme based on object size priors for all remaining objects. 
Therefore, we define for \emph{moving} objects:
\[
     y_i = 
\begin{cases}
    \text{\textit{pedestrian}}, & \text{if} \quad 0.2 < b_w < 1.0~~\text{and}~~0.2 < b_l < 1.0~~\text{and}~~0.8 < b_h < 2.2, \\
    \text{\textit{cyclist}},    & \text{if} \quad 0.2 < b_w < 1.0~~\text{and}~~1.0 < b_l < 2.5~~\text{and}~~1.4 < b_h < 2.0, \\
    \text{\textit{vehicle}}, & \text{if} \quad 0.5 < b_w < 3.0~~\text{and}~~0.5 < b_l < 8.0~~\text{and}~~1.0 < b_h < 3.0, \\
    \text{\textit{background}}, & \text{otherwise.}
\end{cases}
\]
The class label for object $i$ is denoted $y_i$, and the bounding box dimensions \emph{width}, \emph{length}, and \emph{height} are denoted $b_w$, $b_l$, and $b_h$, respectively. 

\paragraph{Temporally-coherent bounding box refinement.}
After propagating median box sizes, we filter those static tracks whose corrected box dimensions deviate significantly from the dimensions of the object categories involved. Therefore, we define the bounding box size thresholds for \emph{width}, \emph{length} and \emph{height} as $0.2 < b_w < 3.5$, $0.2 < b_l < 20.0$ and $0.5 < b_h < 4.0$ respectively.
Finally, to reduce annotation bias, we inflate bounding boxes similar to~\cite{seidenschwanz_2024_semoli} for each dimension by $0.3\si{m}$.

\section{Additional Results}
\label{sec:ablations}
\paragraph{Spatio-temporal clustering.} 
In order to fully exploit the inherent temporal information contained in sequential LiDAR scans, we perform spatio-temporal clustering on multiple LiDAR scans, transformed into the same reference coordinate system.
In \tabref{tab:clustering}, we show the advantage of the proposed spatio-temporal clustering compared to simple frame-by-frame spatial clustering with only spatial input features (x, y, z). 

\begin{table}[h!]
    \small
    \begin{center}
        \begin{tabular}{lccccc}
            \toprule
            \multirow{2}{*}{Clustering}  & AP~(L2) & AP~(L2) & APH~(L2) & APH~(L2) \\
             & BEV & 3D & BEV & 3D \\
            \midrule
            Spatial & 0.351 & 0.306 & 0.250 & 0.212 \\
            Spatio-temporal & 0.363 & 0.323 & 0.260 & 0.225 &    \\
            \bottomrule
        \end{tabular}
    \end{center}
	\caption{
        \textbf{Comparison of spatial and spatio-temporal clustering} following the protocols of~\cite{najibi_2022_motioninspired, baur_2024_liso} on the WOD~\cite{sun_2020_waymo} (\ie AP for BEV and 3D, difficulty level L2, IoU 0.4). We additionally report APH which includes the heading angle precision. 
	}
	\label{tab:clustering}
\end{table}

\paragraph{Size prior baseline comparison.}
Since no comparable approach performs unsupervised class-aware object detection, we implement a baseline classifying objects based on simple size priors.
Therefore, we adopt the default classification scheme for moving objects of our temporally-coherent class label refinement (recall \secref{sec:implementation}) for all objects independent of their motion state. 
Hence, we replace CLIP classification within our approach with simple object size prior thresholds and keep all other parts as is. 
In \tabref{tab:ablation_size_prior}, we show the baseline results compared to our vision-language-guided approach.
We can show that the CLIP model's rich knowledge adds significant value to classifying objects in 3D LiDAR point clouds. 
\begin{table}[ht!]
    \small
    \begin{center}
        \begin{tabular}{lcccccc}
            \toprule
            Classification & \multicolumn{2}{c}{Movable} & Vehicle & Pedestrian &  Cyclist\\
    		method & BEV & 3D & BEV & BEV & BEV \\
            \midrule
            Baseline (size prior) & 0.127 & 0.109 & 0.141 & 0.064 & 0.020 \\
            \OurMethod & 0.363 & 0.323 & 0.490 & 0.168 & 0.076 \\
            \bottomrule
        \end{tabular}
    \end{center}
	
	\caption{
        \textbf{Size prior baseline comparison} following the protocols of~\cite{najibi_2022_motioninspired, baur_2024_liso} on the WOD~\cite{sun_2020_waymo} (\ie AP for BEV and 3D, difficulty level L2, IoU 0.4). We report class-aware detection results for BEV. \OurMethod significantly outperforms the baseline which classifies objects solely on pre-defined object size thresholds.
	}
	\label{tab:ablation_size_prior}
\end{table}

\paragraph{Range evaluation.} 
For the sake of completeness and to gain even more insight, we show the full range evaluation for the Waymo Open Dataset (WOD)~\cite{sun_2020_waymo}.
We provide a detailed range analysis for the zero-shot detection of \OurMethod and the pseudo-label trained Centerpoint~\cite{yin_2021_centerbased} (\OurMethodPlus) in \tabref{tab:zero_shot_range}. 
Following~\cite{sun_2020_waymo, baur_2024_liso}, we report AP at difficulty level L2 with an intersection over union (IoU) threshold
of 0.4 for BEV.

\begin{table}[h]
    \small
    \begin{center}
        \begin{tabular}{llcccc}						
            \toprule			
            \multirow{2}{*}{Method} & \multirow{2}{*}{Class} & Overall & [$0m, 30m$) &  [${30m, 50m}$) & [$50m, +inf)$             \\
            & &  AP / APH & AP / APH & AP / APH & AP / APH \\
            \midrule
            \OurMethod & \multirow{2}{*}{Vehicle} &  0.272 / 0.170 &  0.562 / 0.345 &  0.200 / 0.130 &  0.060 / 0.040 \\
            \OurMethodPlus & &  0.295 / 0.214 &  0.688 / 0.492 &  0.213 / 0.205 &  0.011 / 0.009 \\
            \midrule
            \OurMethod & \multirow{2}{*}{Pedestrian} &  0.123 / 0.113 &  0.200 / 0.180 &  0.090 / 0.085 &  0.060 / 0.058 \\
            \OurMethodPlus & & 0.239 / 0.190 &  0.420 / 0.336 &  0.205 / 0.166 & 0.018 / 0.015 \\
            \midrule
            \OurMethod & \multirow{2}{*}{Cyclist} &  0.047 / 0.047 &  0.080 / 0.079 &  0.025 / 0.025 &  0.019 / 0.019 \\
            \OurMethodPlus & &  0.046 / 0.044 &  0.109 / 0.105 &  0.002 / 0.002 & 0.000 / 0.000 \\
            \bottomrule			
        \end{tabular}
    \end{center}
    \caption{\textbf{Range evaluation} following the protocols of~\cite{sun_2020_waymo, baur_2024_liso} on the WOD~\cite{sun_2020_waymo} (\ie AP for BEV and 3D, difficulty level L2, IoU 0.4). We extend the range to $160 x 160$ around the ego-vehicle. Even in far ranges ($+50\si{m}$), \OurMethod detects some objects correctly.
    }
    \label{tab:zero_shot_range}
\end{table}

\noindent We observe that the detection with \OurMethod and \OurMethodPlus works best in the near range for all object classes. The simple self-supervision with pseudo-labels reinforces the learning for these objects but also improves \emph{pedestrians} and \emph{vehicles} in the medium range by a large margin. 
Only \emph{cyclists}, which are underrepresented in the dataset in the first place and additionally not well detected by \OurMethod, degenerate in the middle to far distances.
Additional augmentations, such as a pseudo-ground truth database or a more complex training routine, could alleviate this negative effect.

\section{Impact of Text Prompts}
\label{sec:text}
\paragraph{Text prompt templates.}
To bridge the modality gap between 3D point clouds and 2D images, we generate depth maps from 3D point segments with varying densities (depending on the distance to the ego-vehicle).
Although not specifically trained for depth images, CLIP~\cite{radford_2021_clip} can still classify many of these projections correctly. 
An important design decision is the text prompt we provide CLIP to get the best feature representation matching the image features. In Table~\ref{tab:text_templates}, we show two additional template variants,~\ie \textit{a depth map of <class>} and \textit{a silhouette of <class>}, describing the projected image. 
It can be observed that \textit{a point representation of <class>} leads to the best results.
However, \textit{a silhouette of <class>} seems to be preferable for \emph{pedestrians} but performs worse overall. 

\begin{table}
    \small
	\begin{center}
    	\begin{tabular}{lcccccc}						
    		\toprule			
    		\multirow{2}{*}{Text prompt template} & \multicolumn{2}{c}{Movable} & Vehicle & Pedestrian &  Cyclist            \\
    		& BEV & 3D & BEV & BEV & BEV \\
    		\midrule
            \textit{a point representation of a <class>} & 0.363 & 0.323 & 0.490 & 0.168 & 0.076 \\
            \midrule
            \textit{a silhouette of a <class>} &  0.258 & 0.223 &  0.294 & 0.190 &  0.075 & \\
        	  \textit{a depth map of <class>} &  0.295 & 0.260 &  0.390 & 0.165 &  0.075 & \\
    		\bottomrule			
    	\end{tabular}
    \end{center}
 	\caption{
          \textbf{Text prompt template evaluation} on WOD~\cite{sun_2020_waymo} (following the protocols of~\cite{najibi_2022_motioninspired, baur_2024_liso}, \ie AP for BEV and 3D, difficulty level L2, IoU 0.4). We show detection results with different text input templates.}
  \label{tab:text_templates}
\end{table}

\bibliography{egbib}

\begin{thebibliography}{74}
\providecommand{\natexlab}[1]{#1}
\providecommand{\url}[1]{\texttt{#1}}
\expandafter\ifx\csname urlstyle\endcsname\relax
  \providecommand{\doi}[1]{doi: #1}\else
  \providecommand{\doi}{doi: \begingroup \urlstyle{rm}\Url}\fi

\bibitem[Barnes et~al.(2018)Barnes, Maddern, Pascoe, and
  Posner]{barnes2018ephemeral}
Dan Barnes, Will Maddern, Geoffrey Pascoe, and Ingmar Posner.
\newblock {Driven to Distraction: Self-Supervised Distractor Learning for
  Robust Monocular Visual Odometry in Urban Environments}.
\newblock In \emph{Proc. ICRA}, 2018.

\bibitem[Baur et~al.(2024)Baur, Moosmann, and Geiger]{baur_2024_liso}
Stefan Baur, Frank Moosmann, and Andreas Geiger.
\newblock {LISO: Lidar-only Self-Supervised 3D Object Detection}.
\newblock \emph{arXiv CoRR}, abs/2403.07071, 2024.

\bibitem[Boulch et~al.(2023)Boulch, Sautier, Michele, Puy, and
  Marlet]{boulch_2022_also}
Alexandre Boulch, Corentin Sautier, Bj{\"o}rn Michele, Gilles Puy, and Renaud
  Marlet.
\newblock {{ALSO}}: {{Automotive Lidar Self-supervision}} by {{Occupancy}}
  estimation.
\newblock In \emph{Proc. CVPR}, 2023.

\bibitem[Caesar et~al.(2020)Caesar, Bankiti, Lang, Vora, Liong, Xu, Krishnan,
  Pan, Baldan, and Beijbom]{caesar_2020_nuscenes}
Holger Caesar, Varun Bankiti, Alex~H. Lang, Sourabh Vora, Venice~Erin Liong,
  Qiang Xu, Anush Krishnan, Yu~Pan, Giancarlo Baldan, and Oscar Beijbom.
\newblock {nuScenes: A multimodal dataset for autonomous driving}.
\newblock In \emph{Proc. CVPR}, 2020.

\bibitem[Caine et~al.(2021)Caine, Roelofs, Vasudevan, Ngiam, Chai, Chen, and
  Shlens]{caine_2021_pseudo}
Benjamin Caine, Rebecca Roelofs, Vijay Vasudevan, Jiquan Ngiam, Yuning Chai,
  Z.~Chen, and Jonathon Shlens.
\newblock {Pseudo-labeling for Scalable 3D Object Detection}.
\newblock \emph{arXiv CoRR}, abs/2103.02093, 2021.

\bibitem[Campello et~al.(2013)Campello, Moulavi, Zimek, and
  Sander]{campello_2013_hdbscan}
Ricardo J. G.~B. Campello, Davoud Moulavi, Arthur Zimek, and J\"{o}rg Sander.
\newblock {Hierarchical Density Estimates for Data Clustering, Visualization,
  and Outlier Detection}.
\newblock In \emph{Proc. PAKDD}, 2013.

\bibitem[Caron et~al.(2020)Caron, Misra, Mairal, Goyal, Bojanowski, and
  Joulin]{caron_2020_swav}
Mathilde Caron, Ishan Misra, Julien Mairal, Priya Goyal, Piotr Bojanowski, and
  Armand Joulin.
\newblock Unsupervised learning of visual features by contrasting cluster
  assignments.
\newblock In \emph{Proc. NeurIPS}, 2020.

\bibitem[Chen et~al.(2023)Chen, Liu, Kong, Zhu, Ma, Li, Hou, Qiao, and
  Wang]{chen_2023_clip_2_scene}
Runnan Chen, Youquan Liu, Lingdong Kong, Xinge Zhu, Yuexin Ma, Yikang Li,
  Yuenan Hou, Yu~Qiao, and Wenping Wang.
\newblock {CLIP2Scene: Towards Label-efficient 3D Scene Understanding by CLIP}.
\newblock In \emph{Proc. CVPR}, 2023.

\bibitem[Deng and Zakhor(2023)]{deng_2023_rsf}
David Deng and Avideh Zakhor.
\newblock {RSF: Optimizing Rigid Scene Flow From 3D Point Clouds Without
  Labels}.
\newblock In \emph{Proc. WACV}, 2023.

\bibitem[Dewan et~al.(2016)Dewan, Caselitz, Tipaldi, and
  Burgard]{dewan_2016_icra}
Ayush Dewan, Tim Caselitz, Gian~Diego Tipaldi, and Wolfram Burgard.
\newblock {Motion-based detection and tracking in 3D LiDAR scans}.
\newblock In \emph{Proc. ICRA}, 2016.

\bibitem[Ding et~al.(2023)Ding, Yang, Xue, Zhang, Bai, and
  Qi]{ding2022language}
Runyu Ding, Jihan Yang, Chuhui Xue, Wenqing Zhang, Song Bai, and Xiaojuan Qi.
\newblock {PLA: Language-Driven Open-Vocabulary 3D Scene Understanding}.
\newblock In \emph{Proc. CVPR}, 2023.

\bibitem[Dosovitskiy et~al.(2021)Dosovitskiy, Beyer, Kolesnikov, Weissenborn,
  Zhai, Unterthiner, Dehghani, Minderer, Heigold, Gelly, Uszkoreit, and
  Houlsby]{dosovitskiy_2021_vitb}
Alexey Dosovitskiy, Lucas Beyer, Alexander Kolesnikov, Dirk Weissenborn,
  Xiaohua Zhai, Thomas Unterthiner, Mostafa Dehghani, Matthias Minderer, Georg
  Heigold, Sylvain Gelly, Jakob Uszkoreit, and Neil Houlsby.
\newblock {An Image is Worth 16x16 Words: Transformers for Image Recognition at
  Scale}.
\newblock In \emph{Proc. ICLR}, 2021.

\bibitem[Fan et~al.(2022)Fan, Pang, Zhang, Wang, Zhao, Wang, Wang, and
  Zhang]{fan_2022_sst}
Lue Fan, Ziqi Pang, Tianyuan Zhang, Yu-Xiong Wang, Hang Zhao, Feng Wang, Naiyan
  Wang, and Zhaoxiang Zhang.
\newblock {Embracing Single Stride 3D Object Detector with Sparse Transformer}.
\newblock In \emph{Proc. CVPR}, 2022.

\bibitem[Fischler and Bolles(1981)]{fischler_1981_ransac}
Martin~A. Fischler and Robert~C. Bolles.
\newblock {Random Sample Consensus: A Paradigm for Model Fitting with
  Applications to Image Analysis and Automated Cartography}.
\newblock \emph{Commun. ACM}, 24\penalty0 (6):\penalty0 381–395, 1981.

\bibitem[Geiger et~al.(2012)Geiger, Lenz, and Urtasun]{geiger_2012_kitti}
Andreas Geiger, Philip Lenz, and Raquel Urtasun.
\newblock {Are we ready for autonomous driving? The KITTI vision benchmark
  suite}.
\newblock In \emph{Proc. CVPR}, 2012.

\bibitem[Himmelsbach et~al.(2010)Himmelsbach, Hundelshausen, and
  Wuensche]{himmelsbach_2010_fastsegment}
M.~Himmelsbach, Felix~v. Hundelshausen, and H.-J. Wuensche.
\newblock {Fast segmentation of 3D point clouds for ground vehicles}.
\newblock In \emph{Proc. IV}, 2010.

\bibitem[Huang et~al.(2024{\natexlab{a}})Huang, Pan, Zheng, Jiang, Xie, Wu,
  Song, and Huang]{huang2023joint}
Rui Huang, Xuran Pan, Henry Zheng, Haojun Jiang, Zhifeng Xie, Cheng Wu, Shiji
  Song, and Gao Huang.
\newblock {Joint representation learning for text and 3D point cloud}.
\newblock \emph{Pattern Recognition}, 147\penalty0 (C):\penalty0 110086,
  2024{\natexlab{a}}.

\bibitem[Huang et~al.(2021)Huang, Xie, Zhu, and
  Zhu]{huang_2021_spatiotemporalself}
Siyuan Huang, Yichen Xie, Song-Chun Zhu, and Yixin Zhu.
\newblock Spatio-{{Temporal Self-Supervised Representation Learning}} for {{3D
  Point Clouds}}.
\newblock In \emph{Proc. ICCV}, 2021.

\bibitem[Huang et~al.(2023)Huang, Dong, Yang, Huang, Lau, Ouyang, and
  Zuo]{huang2022clip2point}
Tianyu Huang, Bowen Dong, Yunhan Yang, Xiaoshui Huang, Rynson~WH Lau, Wanli
  Ouyang, and Wangmeng Zuo.
\newblock {Clip2point: Transfer clip to point cloud classification with
  image-depth pre-training}.
\newblock In \emph{Proc. ICCV}, 2023.

\bibitem[Huang et~al.(2022)Huang, Wang, Guizilini, Ambrus, Gaidon, and
  Solomon]{huang_2022_representation}
Xiangru Huang, Yue Wang, Vitor~Campagnolo Guizilini, Rares~Andrei Ambrus,
  Adrien Gaidon, and Justin Solomon.
\newblock {Representation Learning for Object Detection from Unlabeled Point
  Cloud Sequences}.
\newblock In \emph{Proc. CoRL}, 2022.

\bibitem[Huang et~al.(2024{\natexlab{b}})Huang, Li, Qu, He, Zuo, and
  Ouyang]{huang2022frozen}
Xiaoshui Huang, Sheng Li, Wentao Qu, Tong He, Yifan Zuo, and Wanli Ouyang.
\newblock {EPCL: Frozen CLIP Transformer is An Efficient Point Cloud Encoder}.
\newblock In \emph{Proc. AAAI}, 2024{\natexlab{b}}.

\bibitem[Kingma and Ba(2015)]{diderik_2015_adam}
Diederik~P. Kingma and Jimmy Ba.
\newblock {Adam: A Method for Stochastic Optimization}.
\newblock In \emph{Proc. ICLR}, 2015.

\bibitem[Krispel et~al.(2024)Krispel, Schinagl, Fruhwirth-Reisinger, Possegger,
  and Bischof]{krispel_2023_maeli}
Georg Krispel, David Schinagl, Christian Fruhwirth-Reisinger, Horst Possegger,
  and Horst Bischof.
\newblock {MAELi - Masked Autoencoder for Large-Scale LiDAR Point Clouds}.
\newblock In \emph{Proc. WACV}, 2024.

\bibitem[Lang et~al.(2019)Lang, Vora, Caesar, Zhou, Yang, and
  Beijbom]{lang_2019_pointpillars}
Alex~H. Lang, Sourabh Vora, Holger Caesar, Lubing Zhou, Jiong Yang, and Oscar
  Beijbom.
\newblock {PointPillars: Fast Encoders for Object Detection from Point Clouds}.
\newblock In \emph{Proc. CVPR}, 2019.

\bibitem[Lee et~al.(2018)Lee, Walsh, Harakeh, and
  Waslander]{lee_2018_annotator}
Jungwook Lee, Sean Walsh, Ali Harakeh, and Steven~L. Waslander.
\newblock {Leveraging Pre-Trained 3D Object Detection Models for Fast Ground
  Truth Generation}.
\newblock In \emph{Proc. ITSC}, 2018.

\bibitem[Lee et~al.(2022)Lee, Lim, and Myung]{lee_2022_patchworkpp}
Seungjae Lee, Hyungtae Lim, and Hyun Myung.
\newblock {Patchwork++: Fast and robust ground segmentation solving partial
  under-segmentation using 3D point cloud}.
\newblock In \emph{Proc. IROS}, 2022.

\bibitem[Liang et~al.(2021)Liang, Jiang, Feng, Chen, Xu, Liang, Zhang, Li, and
  Van~Gool]{liang_2021_exploringgeometry}
Hanxue Liang, Chenhan Jiang, Dapeng Feng, Xin Chen, Hang Xu, Xiaodan Liang, Wei
  Zhang, Zhenguo Li, and Luc Van~Gool.
\newblock Exploring {{Geometry-Aware Contrast}} and {{Clustering
  Harmonization}} for {{Self-Supervised 3D Object Detection}}.
\newblock In \emph{Proc. ICCV}, 2021.

\bibitem[Liu et~al.(2023)Liu, Xiao, Zhang, Lu, and Shao]{liu_2023_fac}
Kangcheng Liu, Aoran Xiao, Xiaoqin Zhang, Shijian Lu, and Ling Shao.
\newblock {FAC: 3D Representation Learning via Foreground Aware Feature
  Contrast}.
\newblock In \emph{Proc. CVPR}, 2023.

\bibitem[Lu et~al.(2023)Lu, Xu, Wei, Xie, Tomizuka, Keutzer, and
  Zhang]{lu_2023_openvocabulary}
Yuheng Lu, Chenfeng Xu, Xiaobao Wei, Xiaodong Xie, Masayoshi Tomizuka, Kurt
  Keutzer, and Shanghang Zhang.
\newblock {Open-Vocabulary Point-Cloud Object Detection without 3D Annotation}.
\newblock In \emph{Proc. CVPR}, 2023.

\bibitem[Luo et~al.(2023)Luo, Liu, Chen, You, Benaim, Phoo, Campbell, Sun,
  Hariharan, and Weinberger]{luo_2023_reward}
Katie~Z Luo, Zhenzhen Liu, Xiangyu Chen, Yurong You, Sagie Benaim, Cheng~Perng
  Phoo, Mark Campbell, Wen Sun, Bharath Hariharan, and Kilian~Q Weinberger.
\newblock {Reward Finetuning for Faster and More Accurate Unsupervised Object
  Discovery}.
\newblock In \emph{Proc. NeurIPS}, 2023.

\bibitem[Mao et~al.(2021)Mao, Niu, Jiang, Liang, Li, Ye, Zhang, Li, Yu, and
  Xu]{mao_2021_once}
Jiageng Mao, Minzhe Niu, Chenhan Jiang, Xiaodan Liang, Yamin Li, Chaoqiang Ye,
  Wei Zhang, Zhenguo Li, Jie Yu, and Chunjing Xu.
\newblock {One Million Scenes for Autonomous Driving: ONCE Dataset}.
\newblock In \emph{Proc. NeurIPS}, 2021.

\bibitem[Meng et~al.(2020)Meng, Wang, Zhou, Shen, Van~Gool, and
  Dai]{meng_2020_ws3d}
Qinghao Meng, Wenguan Wang, Tianfei Zhou, Jianbing Shen, Luc Van~Gool, and
  Dengxin Dai.
\newblock {Weakly Supervised 3D Object Detection from Lidar Point Cloud}.
\newblock In \emph{Proc. ECCV}, 2020.

\bibitem[Meng et~al.(2022)Meng, Wang, Zhou, Shen, Jia, and
  Van~Gool]{meng_2022_w3d}
Qinghao Meng, Wenguan Wang, Tianfei Zhou, Jianbing Shen, Yunde Jia, and Luc
  Van~Gool.
\newblock {Towards a Weakly Supervised Framework for 3D Point Cloud Object
  Detection and Annotation}.
\newblock \emph{IEEE TPAMI}, 44\penalty0 (8):\penalty0 4454--4468, 2022.

\bibitem[Min et~al.(2022)Min, Xu, Zhao, Xiao, Nie, and Dai]{min_2022_voxelmae}
Chen Min, Xinli Xu, Dawei Zhao, Liang Xiao, Yiming Nie, and Bin Dai.
\newblock Voxel-{{MAE}}: {{Masked Autoencoders}} for {{Pre-training Large-scale
  Point Clouds}}.
\newblock \emph{arXiv CoRR}, abs/2206.09900, 2022.

\bibitem[Najibi et~al.(2022)Najibi, Ji, Zhou, Qi, Yan, Ettinger, and
  Anguelov]{najibi_2022_motioninspired}
Mahyar Najibi, Jingwei Ji, Yin Zhou, Charles~R. Qi, Xinchen Yan, Scott
  Ettinger, and Dragomir Anguelov.
\newblock {Motion Inspired Unsupervised Perception and Prediction in Autonomous
  Driving}.
\newblock In \emph{Proc. CVPR}, 2022.

\bibitem[Peng et~al.(2023)Peng, Genova, Jiang, Tagliasacchi, Pollefeys, and
  Funkhouser]{peng_2023_openscene}
Songyou Peng, Kyle Genova, Chiyu~"Max" Jiang, Andrea Tagliasacchi, Marc
  Pollefeys, and Thomas Funkhouser.
\newblock {OpenScene: 3D Scene Understanding with Open Vocabularies}.
\newblock In \emph{Proc. CVPR}, 2023.

\bibitem[Qi et~al.(2021)Qi, Zhou, Najibi, Sun, Vo, Deng, and
  Anguelov]{qi_2021_offboard}
Charles~R. Qi, Yin Zhou, Mahyar Najibi, Pei Sun, Khoa Vo, Boyang Deng, and
  Dragomir Anguelov.
\newblock {Offboard 3D Object Detection From Point Cloud Sequences}.
\newblock In \emph{Proc. CVPR}, 2021.

\bibitem[Radford et~al.(2021)Radford, Wook~Kim, Hallacy, Ramesh, Goh, Agarwal,
  Sastry, Askell, Mishkin, Clark, Krueger, and Sutskever]{radford_2021_clip}
Alec Radford, Jong Wook~Kim, Chris Hallacy, Aditya Ramesh, Gabriel Goh,
  Sandhini Agarwal, Girish Sastry, Amanda Askell, Pamela Mishkin, Jack Clark,
  Gretchen Krueger, and Ilya Sutskever.
\newblock {Learning Transferable Visual Models From Natural Language
  Supervision}.
\newblock In \emph{Proc. ICML}, 2021.

\bibitem[Seidenschwarz et~al.(2024)Seidenschwarz, Ošep, Ferroni, Lucey, and
  Leal-Taixé]{seidenschwanz_2024_semoli}
Jenny Seidenschwarz, Aljoša Ošep, Francesco Ferroni, Simon Lucey, and Laura
  Leal-Taixé.
\newblock {SeMoLi: What Moves Together Belongs Together}.
\newblock In \emph{Proc. CVPR}, 2024.

\bibitem[Shi et~al.(2022)Shi, Li, and Ma]{shi_2022_pillarnet}
Guangsheng Shi, Ruifeng Li, and Chao Ma.
\newblock {PillarNet: High-Performance Pillar-based 3D Object Detection}.
\newblock In \emph{Proc. ECCV}, 2022.

\bibitem[Shi et~al.(2019)Shi, Wang, and Li]{shi_2019_pointrcnn}
Shaoshuai Shi, Xiaogang Wang, and Hongsheng Li.
\newblock {PointRCNN: 3D Object Proposal Generation and Detection from Point
  Cloud}.
\newblock In \emph{Proc. CVPR}, 2019.

\bibitem[Shi et~al.(2020)Shi, Guo, Jiang, Wang, Shi, Wang, and
  Li]{shi_2020_pvrcnn}
Shaoshuai Shi, Chaoxu Guo, Li~Jiang, Zhe Wang, Jianping Shi, Xiaogang Wang, and
  Hongsheng Li.
\newblock {PV-RCNN: Point-Voxel Feature Set Abstraction for 3D Object
  Detection}.
\newblock In \emph{Proc. CVPR}, 2020.

\bibitem[Shi et~al.(2023)Shi, Jiang, Deng, Wang, Guo, Shi, Wang, and
  Li]{shi_2021_pvrcnnpp}
Shaoshuai Shi, Li~Jiang, Jiajun Deng, Zhe Wang, Chaoxu Guo, Jianping Shi,
  Xiaogang Wang, and Hongsheng Li.
\newblock {PV-RCNN++: Point-Voxel Feature Set Abstraction With Local Vector
  Representation for 3D Object Detection}.
\newblock \emph{IJCV}, 131\penalty0 (6):\penalty0 531–551, 2023.

\bibitem[Shi and Rajkumar(2020)]{shi_2020_pointgnn}
Weijing Shi and Ragunathan Rajkumar.
\newblock {Point-GNN: Graph Neural Network for 3D Object Detection in a Point
  Cloud}.
\newblock In \emph{Proc. CVPR}, 2020.

\bibitem[Smith(2018)]{smith_2018_onecycle}
Leslie~N. Smith.
\newblock {A Disciplined Approach to Neural Network Hyper-Parameters: {{Part}}
  1 -- Learning Rate, Batch Size, Momentum, and Weight Decay}.
\newblock \emph{arXiv CoRR}, abs/1803.09820, 2018.

\bibitem[Sualeh and Kim(2019)]{sualeh_2019_dynamic}
Muhammad Sualeh and Gon-Woo Kim.
\newblock {Dynamic Multi-LiDAR Based Multiple Object Detection and Tracking}.
\newblock \emph{Sensors}, 19\penalty0 (10):\penalty0 1474, 2019.

\bibitem[Sun et~al.(2020)Sun, Kretzschmar, Dotiwalla, Chouard, Patnaik, Tsui,
  Guo, Zhou, Chai, Caine, et~al.]{sun_2020_waymo}
Pei Sun, Henrik Kretzschmar, Xerxes Dotiwalla, Aurelien Chouard, Vijaysai
  Patnaik, Paul Tsui, James Guo, Yin Zhou, Yuning Chai, Benjamin Caine, et~al.
\newblock {Scalability in Perception for Autonomous Driving: Waymo Open
  Dataset}.
\newblock In \emph{Proc. CVPR}, 2020.

\bibitem[Tang and Lee(2019)]{tang_2019_transferable3D}
Yew~Siang Tang and Gim~Hee Lee.
\newblock {Transferable Semi-supervised 3D Object Detection from RGB-D Data}.
\newblock In \emph{Proc. ICCV}, 2019.

\bibitem[Team(2020)]{pcdetteam_2020_openpcdet}
OpenPCDet~Development Team.
\newblock {OpenPCDet}: An open-source toolbox for {3D} object detection from
  point clouds.
\newblock \url{https://github.com/open-mmlab/OpenPCDet}, 2020.

\bibitem[Wang et~al.(2020)Wang, Chen, You, Erran, Hariharan, Campbell,
  Weinberger, and Chao]{wang_2020_train}
Yan Wang, Xiangyu Chen, Yurong You, Li~Erran, Bharath Hariharan, Mark Campbell,
  Kilian~Q. Weinberger, and Wei-Lun Chao.
\newblock {Train in germany, test in the usa: Making 3d object detectors
  generalize}.
\newblock In \emph{Proc. CVPR}, 2020.

\bibitem[Wang et~al.(2022)Wang, Chen, and Zhang]{wang_2022_4d}
Yuqi Wang, Yuntao Chen, and Zhaoxiang Zhang.
\newblock {4D Unsupervised Object Discovery}.
\newblock In \emph{Proc. NeurIPS}, 2022.

\bibitem[Wilson et~al.(2021)Wilson, Qi, Agarwal, Lambert, Singh, Khandelwal,
  Pan, Kumar, Hartnett, Pontes, Ramanan, Carr, and
  Hays]{wilson_2021_argoverse2}
Benjamin Wilson, William Qi, Tanmay Agarwal, John Lambert, Jagjeet Singh,
  Siddhesh Khandelwal, Bowen Pan, Ratnesh Kumar, Andrew Hartnett,
  Jhony~Kaesemodel Pontes, Deva Ramanan, Peter Carr, and James Hays.
\newblock {Argoverse 2: Next Generation Datasets for Self-driving Perception
  and Forecasting}.
\newblock In \emph{Proc. NeurIPS}, 2021.

\bibitem[Wu et~al.(2015)Wu, Song, Khosla, Yu, Zhang, Tang, and Xiao]{wu20153d}
Zhirong Wu, Shuran Song, Aditya Khosla, Fisher Yu, Linguang Zhang, Xiaoou Tang,
  and Jianxiong Xiao.
\newblock {3d shapenets: A deep representation for volumetric shapes}.
\newblock In \emph{Proc. CVPR}, pages 1912--1920, 2015.

\bibitem[Xu et~al.(2023)Xu, Wang, Zhang, Chen, Cao, Pang, and
  Lin]{xu_2023_mvjar}
Runsen Xu, Tai Wang, Wenwei Zhang, Runjian Chen, Jinkun Cao, Jiangmiao Pang,
  and Dahua Lin.
\newblock {MV-JAR: Masked Voxel Jigsaw and Reconstruction for LiDAR-Based
  Self-Supervised Pre-Training}.
\newblock In \emph{Proc. CVPR}, 2023.

\bibitem[Xue et~al.(2023)Xue, Gao, Xing, Mart{\'\i}n-Mart{\'\i}n, Wu, Xiong,
  Xu, Niebles, and Savarese]{xue2022ulip}
Le~Xue, Mingfei Gao, Chen Xing, Roberto Mart{\'\i}n-Mart{\'\i}n, Jiajun Wu,
  Caiming Xiong, Ran Xu, Juan~Carlos Niebles, and Silvio Savarese.
\newblock Ulip: Learning unified representation of language, image and point
  cloud for 3d understanding.
\newblock In \emph{Proc. CVPR}, 2023.

\bibitem[Yan et~al.(2018)Yan, Mao, and Li]{yan_2018_second}
Yan Yan, Yuxing Mao, and Bo~Li.
\newblock {SECOND: Sparsely Embedded Convolutional Detection}.
\newblock \emph{Sensors}, 18\penalty0 (10):\penalty0 3337, 2018.

\bibitem[Yang et~al.(2021)Yang, Bai, Liang, Zeng, and
  Urtasun]{yang_2021_auto4d}
Bin Yang, Min Bai, Ming Liang, Wenyuan Zeng, and Raquel Urtasun.
\newblock {Auto4D: Learning to Label 4D Objects from Sequential Point Clouds}.
\newblock \emph{arXiv CoRR}, abs/2101.06586, 2021.

\bibitem[Yang et~al.(2020)Yang, Sun, Liu, and Jia]{yang_2020_3dssd}
Zetong Yang, Yanan Sun, Shu Liu, and Jiaya Jia.
\newblock {3DSSD: Point-based 3D Single Stage Object Detector}.
\newblock In \emph{Proc. CVPR}, 2020.

\bibitem[Yao et~al.(2022)Yao, Zhang, Yin, Luo, Ouyang, and Huang]{yao20223d}
Yuan Yao, Yuanhan Zhang, Zhenfei Yin, Jiebo Luo, Wanli Ouyang, and Xiaoshui
  Huang.
\newblock {3D Point Cloud Pre-training with Knowledge Distillation from 2D
  Images}.
\newblock \emph{arXiv CoRR}, abs/2212.08974, 2022.

\bibitem[Yin et~al.(2022{\natexlab{a}})Yin, Fang, Zhou, Zhang, Xu, Shen, and
  Wang]{yin_2022_proficient_teachers}
Junbo Yin, Jin Fang, Dingfu Zhou, Liangjun Zhang, Cheng-Zhong Xu, Jianbing
  Shen, and Wenguan Wang.
\newblock {Semi-supervised 3D Object Detection with Proficient Teachers}.
\newblock In \emph{Proc. ECCV}, 2022{\natexlab{a}}.

\bibitem[Yin et~al.(2022{\natexlab{b}})Yin, Zhou, Zhang, Fang, Xu, Shen, and
  Wang]{yin_2022_proposalcontrast}
Junbo Yin, Dingfu Zhou, Liangjun Zhang, Jin Fang, Cheng-Zhong Xu, Jianbing
  Shen, and Wenguan Wang.
\newblock {{ProposalContrast}}: {{Unsupervised Pre-training}} for {{LiDAR-based
  3D Object Detection}}.
\newblock In \emph{Proc. ECCV}, 2022{\natexlab{b}}.

\bibitem[Yin et~al.(2021)Yin, Zhou, and
  Kr{\"a}henb{\"u}hl]{yin_2021_centerbased}
Tianwei Yin, Xingyi Zhou, and Philipp Kr{\"a}henb{\"u}hl.
\newblock {Center-based 3D Object Detection and Tracking}.
\newblock In \emph{Proc. CVPR}, 2021.

\bibitem[You et~al.(2022{\natexlab{a}})You, Luo, Phoo, Chao, Sun, Hariharan,
  Campbell, and Weinberger]{you_2022_mobiledetection}
Yurong You, Katie~Z Luo, Cheng~Perng Phoo, Wei-Lun Chao, Wen Sun, Bharath
  Hariharan, Mark Campbell, and Kilian~Q. Weinberger.
\newblock {Learning to Detect Mobile Objects from LiDAR Scans Without Labels}.
\newblock In \emph{Proc. CVPR}, 2022{\natexlab{a}}.

\bibitem[You et~al.(2022{\natexlab{b}})You, Phoo, Luo, Zhang, Chao, Hariharan,
  Campbell, and Weinberger]{you_2022_udarepeatedtraversals}
Yurong You, Cheng~Perng Phoo, Katie~Z Luo, Travis Zhang, Wei-Lun Chao, Bharath
  Hariharan, Mark Campbell, and Kilian~Q. Weinberger.
\newblock {Unsupervised Adaptation from Repeated Traversals for Autonomous
  Driving}.
\newblock In \emph{Proc. NeurIPS}, 2022{\natexlab{b}}.

\bibitem[Zakharov et~al.(2020)Zakharov, Kehl, Bhargava, and
  Gaidon]{zakharov_2020_sdflabel}
Sergey Zakharov, Wadim Kehl, Arjun Bhargava, and Adrien Gaidon.
\newblock {Autolabeling 3D Objects with Differentiable Rendering of SDF Shape
  Priors}.
\newblock In \emph{Proc. CVPR}, 2020.

\bibitem[Zeng et~al.(2023)Zeng, Jiang, Mao, Han, Ye, Huang, Yeung, Yang, Liang,
  and Xu]{zeng_2023_clip_square}
Yihan Zeng, Chenhan Jiang, Jiageng Mao, Jianhua Han, Chaoqiang Ye, Qingqiu
  Huang, Dit-Yan Yeung, Zhen Yang, Xiaodan Liang, and Hang Xu.
\newblock {Contrastive Language-Image-Point Pretraining from Real-World Point
  Cloud Data}.
\newblock In \emph{Proc. CVPR}, 2023.

\bibitem[Zhang et~al.(2023)Zhang, Yang, Xiong, Casas, Yang, Ren, and
  Urtasun]{zhang_2023_oyster}
Lunjun Zhang, Anqi~Joyce Yang, Yuwen Xiong, Sergio Casas, Bin Yang, Mengye Ren,
  and Raquel Urtasun.
\newblock {Towards Unsupervised Object Detection From LiDAR Point Clouds}.
\newblock In \emph{Proc. CVPR}, 2023.

\bibitem[Zhang et~al.(2013)Zhang, Song, Shao, Zhao, and
  Shibasaki]{zhang_2013_unsupervised3d}
Quanshi Zhang, Xuan Song, Xiaowei Shao, Huijing Zhao, and Ryosuke Shibasaki.
\newblock {Unsupervised 3D category discovery and point labeling from a large
  urban environment}.
\newblock In \emph{Proc. ICRA}, 2013.

\bibitem[Zhang et~al.(2022{\natexlab{a}})Zhang, Guo, Zhang, Li, Miao, Cui,
  Qiao, Gao, and Li]{zhang_2021_pointclip}
Renrui Zhang, Ziyu Guo, Wei Zhang, Kunchang Li, Xupeng Miao, Bin Cui, Yu~Qiao,
  Peng Gao, and Hongsheng Li.
\newblock {PointCLIP: Point Cloud Understanding by CLIP}.
\newblock In \emph{Proc. CVPR}, 2022{\natexlab{a}}.

\bibitem[Zhang et~al.(2017)Zhang, Xu, Dong, and Dolan]{zhang_2017_lshape}
Xiao Zhang, Wenda Xu, Chiyu Dong, and John~M. Dolan.
\newblock {Efficient L-shape fitting for vehicle detection using laser
  scanners}.
\newblock In \emph{Proc. IV}, 2017.

\bibitem[Zhang et~al.(2022{\natexlab{b}})Zhang, Hu, Xu, Ma, Wan, and
  Guo]{zhang_2022_notallpoints}
Yifan Zhang, Qingyong Hu, Guoquan Xu, Yanxin Ma, Jianwei Wan, and Yulan Guo.
\newblock {Not all points are equal: Learning highly efficient point-based
  detectors for 3d lidar point clouds}.
\newblock In \emph{Proc. CVPR}, 2022{\natexlab{b}}.

\bibitem[Zhou and Tuzel(2018)]{zhou_2018_voxelnet}
Yin Zhou and Oncel Tuzel.
\newblock {VoxelNet: End-to-End Learning for Point Cloud Based 3D Object
  Detection}.
\newblock In \emph{Proc. CVPR}, 2018.

\bibitem[Zhou et~al.(2022)Zhou, Zhao, Wang, Wang, and
  Foroosh]{zhou_2022_centerformer}
Zixiang Zhou, Xiangchen Zhao, Yu~Wang, Panqu Wang, and Hassan Foroosh.
\newblock {CenterFormer: Center-based Transformer for 3D Object Detection}.
\newblock In \emph{Proc. ECCV}, 2022.

\bibitem[Zhu et~al.(2023)Zhu, Zhang, He, Zeng, Zhang, and
  Gao]{zhu_2022_pointclipv2}
Xiangyang Zhu, Renrui Zhang, Bowei He, Ziyao Zeng, Shanghang Zhang, and Peng
  Gao.
\newblock {PointCLIP V2: Prompting CLIP and GPT for Powerful 3D Open-world
  Learning}.
\newblock In \emph{Proc. CVPR}, 2023.

\end{thebibliography}
\end{document}